\documentclass[a4paper]{article}
\usepackage{arxiv}
\usepackage[labelfont=bf]{caption}
\usepackage[utf8]{inputenc}
\usepackage{enumitem}
\usepackage{textcomp}
\usepackage{amsmath}
\usepackage{bm}
\usepackage{amssymb}
\usepackage{stackrel}
\usepackage{graphicx}
\usepackage{wasysym}
\usepackage{esint}
\usepackage{cite}
\usepackage{subcaption}

\usepackage{etoolbox} 

\usepackage[font=small,labelfont=bf]{caption}
\usepackage[unicode=true,
 bookmarks=false,
 breaklinks=false,pdfborder={0 0 1},backref=section,colorlinks=false]
 {hyperref}
\makeatletter
\newcommand{\lyxdot}{.}

\newcommand{\myss}[3]{\sideset{#1}{#2}{\mathop{#3}}}
\makeatother

\begin{document}
\institute{Institute for Flight Mechanics and Controls, University of Stuttgart, Stuttgart, Germany (\email{(eric.price,aamir.ahmad)@ifr.uni-stuttgart.de})\and Max Planck Institute for Intelligent Systems, Tübingen, Germany (\email{mjb@tuebingen.mpg.de})}

\docdate{2023-03-19}

\title{Viewpoint-driven Formation Control of Airships for Cooperative Target Tracking}

\author{Eric Price\inst{1,2}, Michael J. Black\inst{2} and Aamir Ahmad\inst{1,2}
\thanks{We thank Yu Tang Liu, Pascal Goldschmid, Egor Iuganov, Christian Gall
and Ruben Leidel at the Flight Robotic and Perception Group for their
assistance with the flight experiments, as well as Pascal Goldschmid for
extremely helpful discussions.} }\maketitle

\begin{abstract}
For tracking and motion capture (MoCap) of animals in their
natural habitat, a formation of safe and silent aerial platforms, such
as airships with on-board cameras, is well suited. In our prior work we
derived formation properties for optimal MoCap, which include maintaining constant angular separation between observers w.r.t. the subject, threshold distance to it and keeping it centered in the camera view. 
\
Unlike multi-rotors, airships have non-holonomic constrains and are affected by ambient wind. Their orientation and flight direction are also tightly coupled. Therefore a control scheme for multicopters that assumes independence of motion direction and orientation is not applicable. 
\
In this paper, we address this problem by first exploiting a periodic relationship between the airspeed of an airship and its distance to the subject. We use it to derive analytical and numeric solutions that satisfy the formation properties for optimal MoCap. Based on this, we develop an MPC-based formation controller. 
\
We perform theoretical analysis of our solution, boundary conditions
of its applicability, extensive simulation experiments and a real world
demonstration of our control method with an unmanned airship. Open source code
\url{https://tinyurl.com/AsMPCCode} and a video
of our method is provided at \url{https://tinyurl.com/AsMPCVid}.
\end{abstract}

\section{Introduction}

\setlength{\belowcaptionskip}{-15pt}
\begin{figure}[t]
\centering{}\includegraphics[width=1\columnwidth]{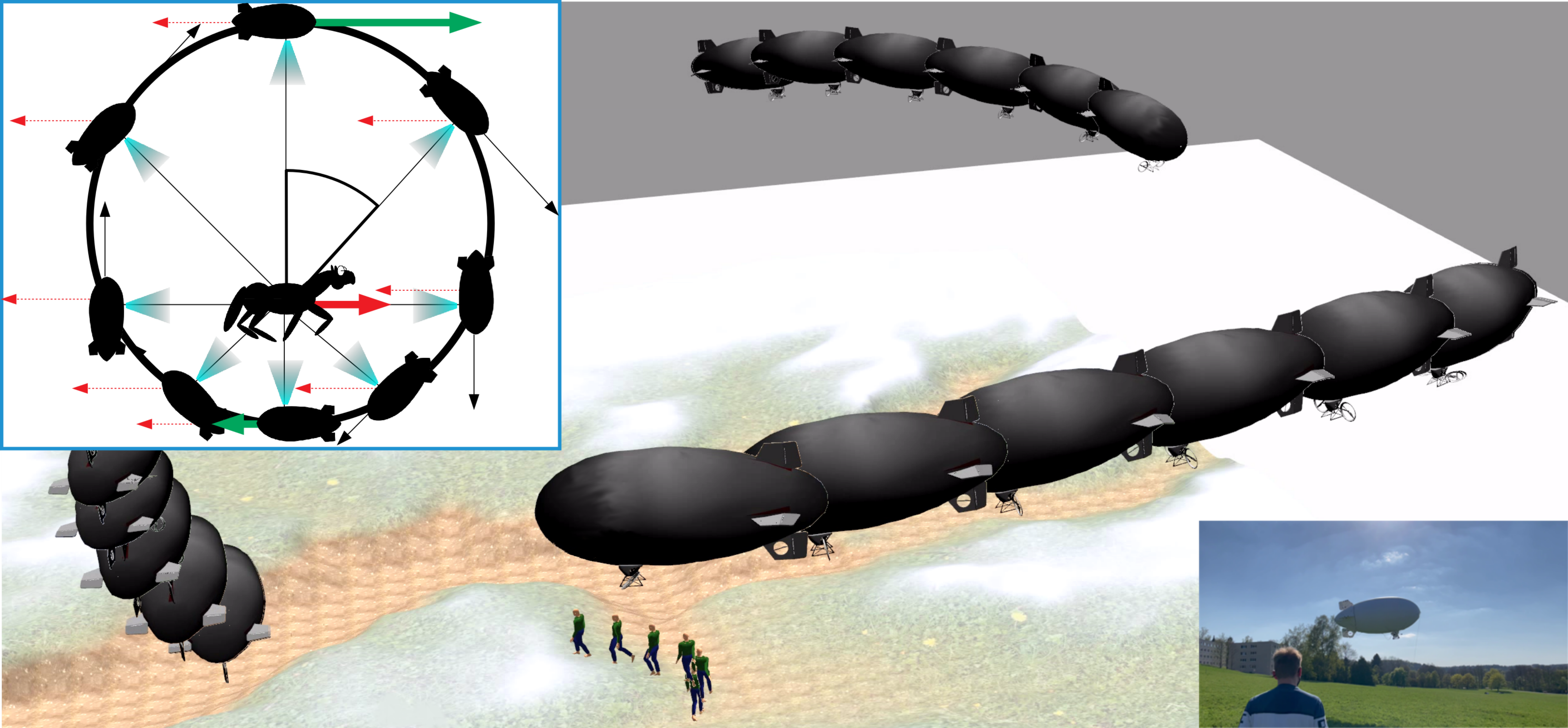}
\caption{\label{fig:orbit}
Multi-exposure visualization of 3 simulated airships in a formation, tracking a moving person. Inset Left: top-view illustration of an 8-blimp formation orbiting a subject. Red arrows: relative motion w.r.t.\ subject due to subject motion, resulting in periodic changes in orbit radius. Black arrows: airspeed. Green arrows: the airspeeds at the furthest and closest points, increased or decreased, respectively, by two times the subject velocity. Inset Right: Our real blimp during an experiment.
}
\end{figure}
\setlength{\belowcaptionskip}{-10pt}

Aerial motion capture (MoCap) of human and animal subjects, i.e., estimating the trajectory of their 3D positions, skeletal poses and body shape, using a team of aerial vehicles with on-board cameras, is a challenging task in various monitoring applications. For optimal aerial MoCap, in our previous work we developed an MPC-based \cite{8784232} and a reinforcement learning-based  \cite{aircaprl} formation control method for multicopters. For human pose and shape estimation from multiple aerial images, our previous works include \cite{Saini:IRL:2022} and \cite{8394622}. In these works, we have shown that observing an articulated subject, like a human, from multiple UAVs avoids self-occlusions in the subject and improves the accuracy of the MoCap estimates. In \cite{8784232} and \cite{aircaprl}, we also established that optimal MoCap estimates can be achieved if i) the subject remains centered in the camera images of all UAVs, ii) the UAVs maintain a threshold distance to the subject within safety limits, and iii) the UAVs maintain a formation-size-specific angular formation around the subject. 

While these formation properties are agnostic to the vehicle-type, the controllers presented in both \cite{8784232} and \cite{8394622} assume multicopter UAVs. These are not well-suited for some subjects, such as animals in their natural habitat, due to their loud noise, low flight time (battery) and safety-related concerns. Buoyant, helium-based, lighter than air vehicles (or airships) \cite{Liu_IROS_22,Price:IAS:2021} can address these concerns. These are relatively safe in terms of collision-related damage. Compared to multi-copters capable of carrying similar payloads, airships can have significantly longer flight times and considerably lower noise levels.

Unfortunately, airship motion and control is fundamentally non-holonomic in nature and substantially different from that of the multicopters. Therefore, the solutions developed in \cite{8784232} and \cite{8394622} are not directly applicable. Most airship designs need to remain in motion to be controllable. Their orientation and flight direction are also tightly coupled. For example, in order to climb, an airship typically needs to pitch up, which rotates the camera view around its lateral Y-axis, assuming a rigidly mounted camera. A directional change, on the other hand, triggers a roll around the airship's longitudinal X-axis. Most importantly, classical airship designs allow movement only in the forward direction through the surrounding air, offset by a lateral angle of attack. While the global camera orientation could be decoupled from flight direction using gimbal-mounted cameras, these add additional weight and control complexity. Therefore, a solution is required for airship formation control for the task of aerial MoCap that i) does not need camera gimbals, and ii) adheres to the previously-mentioned optimal MoCap conditions. This is the core problem addressed in this paper. 

A possible solution for a stationary subject and a single airship, is to mount the camera perpendicular to the forward direction of the airship, tilted downward at an angle, and let the airship orbit around the subject. For a formation of airships, maintaining the same angular velocity by all airships would be a solution. However, for subjects moving with arbitrary and changing velocities, these solutions are not directly applicable. The key contribution of this paper is that we show, both theoretically and empirically, that subject-orbiting trajectories of airships can be generalized even to subjects in motion, while maintaining a perpendicular line-of-sight (LoS) to the subject and a prescribed angular separation between them. We achieve this by exploiting a periodic relationship between the airspeed of the airship and its distance to the subject. We also show how our method copes with wind, as both wind and subject motion result in equivalent relative velocity between the airship and the subject. Our further novel contributions include the following.

\begin{itemize}[leftmargin=*]
 \item We derive analytical solutions for a simplified 2D case (airships and the subject moving on the same plane) with simplified aerodynamics. From this, we derive boundary conditions on the subject velocity that allow possible orbit solutions under the constraints of maximum and minimum airship speeds. 
 \item We then derive an approximate analytical solution in the full 3D case with realistic physics. Based on this, and a MoCap cost function, we develop a numeric solver for the formations. Utilizing this solver, we implement a fixed-time model predictive controller (MPC) for the formation control problem. 
\end{itemize}

We evaluate our MPC in simulation experiments (Fig.~\ref{fig:orbit}) varying parameters such as wind, subject motion, and the number of vehicles in the formation. Finally, we demonstrate it on our real airship. We opened the source code for the benefit of the community.

\section{Related works\label{sec:State-of-the}}

MPC has been extensively used for drone formation control \cite{doi:10.1080/01691864.2018.1470572,8784232,8287413,iros2021}. Most works on airship formation control, with \cite{4108474}
or without \cite{airshipformationcontrol} MPC cover traditional
formation types with pre-specified geometry. However, none 
considers centering a subject in the on-board camera's image
or angular separation constraints with respect to a tracked subject.

Since the non-holonomic motion constraints of airships and fixed-wing aircraft
are similar enough to allow cross-application, we briefly overview some works in that field.
Subject-relative formations have been well studied for observation with fixed-wings, 
but primarily with straight downward facing cameras, where the vehicles loiter high above the
subject \cite{6864849,LIAO2020811,Muslimov2021,Zhang2016}. This greatly
simplifies the formation control problem, while simultaneously limiting
the observation angles to a top down birds-eye view, which is not
well suitable for the MoCap task. In contrast, our approach can be employed
for arbitrary observation angles and for subjects at the same altitude or even higher than the airship.

Orbiting trajectories for fixed-wing UAVs around a subject have been
described in \cite{9290189}, where a Lyapunov function based vector
field is designed to explicitly converge on an orbiting solution in
a constant altitude plane. It maintains angular separation between
vehicles and constant distance to the subject, but it does not consider
vehicle orientation and camera viewing angles relative to the subject.
The Lyapunov function based approach is sensitive to starting conditions
and can not easily be extended to maintain additional constraints.
Our optimization based approach not only converges on steady state
solutions, but also optimizes the transition phases for optimal subject
coverage, considers motion in 3D space and can enforce constraints
on both state and control inputs. Other works \cite{takei2012efficient,doi:10.1137/17M1152589,ZHOU201664} deal with subject-relative formation strategies for varying objectives, but do not consider body-frame limited viewing angles or non-holonomic motion constraints.

\section{Methodology\label{sec:Methodology}}

\subsection{Notations\label{subsec:Notation}}

\begin{figure}[!t]
\begin{centering}
\includegraphics[width=1\columnwidth]{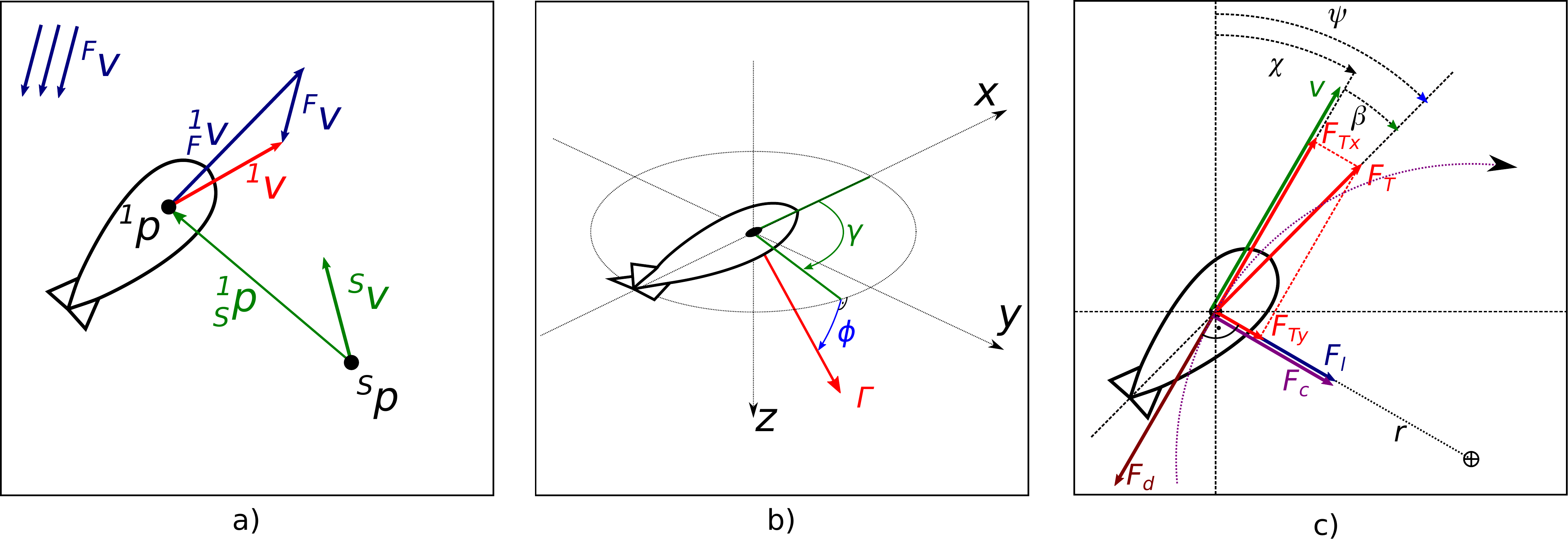}
\par\end{centering}
\caption{\label{fig:Accelerations-on-an}a) Absolute and relative position
and velocity of airship $1$ near subject $S$. b) Camera angle in
the blimp body frame. c) Forces on an airship on a curved trajectory,
subject to aerodynamic lift and drag.}
\end{figure}

In the world frame, the velocity of airship $n\in\left[1\ldots N\right]$
at time $t$ is given as $\myss{^{n}}{_{t}}{\bm{v}}=\left[\myss{^{n}}{_{t}}{v_{x}},\myss{^{n}}{_{t}}{v_{y}},\myss{^{n}}{_{t}}{v_{z}}\right]^{\top}$ and
its horizontal speed is given as $\myss{^{n}}{_{t}}{v_{h}}=\left\Vert \left[\myss{^{n}}{_{t}}{v_{x}},\myss{^{n}}{_{t}}{v_{y}}\right]\right\Vert \mathrm{.}$
The wind/fluid vector is given as $\myss{^{F}}{}{\bm{v}}$. The velocity
of the subject is given as $\myss{^{S}}{}{\bm{v}}$.
The position of airship
$n$ and subject $S$ at time $t$ are given as $\myss{^{n}}{_{t}}{\bm{p}}$
and $\myss{^{S}}{_{t}}{\bm{p}}$, respectively. We use a left subscript
to denote a reference frame different from world frame, e.g., the velocity
of airship $n$ at time $t$ relative to the fluid $F$ is given as $\myss{_{F}^{n}}{_{t}}{\bm{v}}=\myss{^{n}}{_{t}}{\bm{v}}-\myss{^{F}}{}{\bm{v}}$ and the position of airship $n$ relative to subject $S$
is given as $\myss{_{S}^{n}}{_{t}}{\bm{p}}=\myss{^{n}}{_{t}}{\bm{p}}-\myss{^{S}}{_{t}}{\bm{p}}$ (Fig. \ref{fig:Accelerations-on-an}a).

All positions and velocities are oriented North-East-Down, unless specified
otherwise. In airship body frame, we consider $x$ forward, $y$ rightward
and $z$ downward. The orientation of airship $n$ at time $t$ is
given as $\myss{^{n}}{_{t}}{\bm{\Theta}}=\left[\myss{^{n}}{_{t}}{\varphi},\myss{^{n}}{_{t}}{\theta},\myss{^{n}}{_{t}}{\psi}\right]^{\top}$,
with components roll $\myss{^{n}}{_{t}}{\varphi}$, pitch $\myss{^{n}}{_{t}}{\theta}$
and yaw $\myss{^{n}}{_{t}}{\psi}$ as defined in \cite[p. 147]{von1963aerodynamics}. The horizontal motion direction
of airship $n$ in the fluid is given as
\begin{equation}
\myss{_{F}^{n}}{_{t}}{\chi}=\mathrm{atan2}\left(\frac{\myss{_{F}^{n}}{_{t}}{v_{y}}}{\myss{_{F}^{n}}{_{t}}{v_{x}}}\right)\mathrm{.}
\end{equation}

We write $\myss{^{n}}{_{t}}{\dot{\psi}}$ for the yaw rate
and $\myss{_{F}^{n}}{_{t}}{\dot{\chi}}$ for the turn rate of airship
$n$ at time $t$. We define the formation state of $N$ airships,
at time $t$ as
\begin{equation}
\bm{X}_{t}=\left[\myss{^{S}}{_{t}}{\bm{x}},\myss{^{1}}{_{t}}{\bm{x}},\ldots,\myss{^{N}}{_{t}}{\bm{x}}\right]^{\top},
\end{equation}consisting of subject state $\myss{^{S}}{_{t}}{\bm{x}}=\left[\myss{^{S}}{_{t}}{\bm{p}}\right]$ and airship states $\myss{^{n}}{_{t}}{\bm{x}}=\left[\myss{^{n}}{_{t}}{\bm{p}},\myss{^{n}}{_{t}}{\bm{v}},\myss{^{n}}{_{t}}{\bm{\Theta}}\right]^{\top}$ for all airships $n\in\left[1\ldots N\right]$.

\subsection{Problem Statement\label{subsec:Problem-Statement}}

Let us assume a team of $N$ airships, each with a body-fixed camera, tracking a moving subject $S$.
Let us also assume that the camera's fixed azimuth and elevation angles with respect to its carrier airship is given as $\gamma$ and $\phi$, respectively, and represented jointly as $\bm{\Gamma}=\left[\gamma,\phi\right]$.
Our goal is to control the airship's acceleration $\myss{_{F}^{n}}{}{\dot{v}}$ and yaw rate $\myss{^{n}}{_{t}}{\dot{\psi}}$, such that the following conditions are met.
\begin{enumerate}
\item $S$ remains centered in the camera view for all $n\in\left[1\ldots N\right]$, a MoCap accuracy specific requirement \cite{aircaprl}.
\item The angles subtended by any two airships $n,m$, with respect to the subject $S$ remain equal to a defined preset that is optimal for minimizing the joint state estimation uncertainty. In \cite{8784232}, it was shown that an angle of $\frac{2\pi}{N}$ for $N>2$ and $\frac{\pi}{2}$ for $N=2$
is optimal.
\item Other input and state constraints: we assume that in order to remain maneuverable, each airship must fly above a minimum airspeed $v_{\min}$ and, due to drag and finite thrust, cannot exceed $v_{\max}$ such that
\begin{equation}
v_{\min}\leq\left\Vert \myss{_{F}^{n}}{_{t}}{\bm{v}}\right\Vert \leq v_{\max}\mathrm{.}\label{eq:2}
\end{equation}
\end{enumerate}

\subsection{Planar Airship Orbits in 2D\label{subsec:Analytic-solution-for}}

For our primary analysis, we consider a system with the following
simplifications: i) movement and rotation in a 2D X/Y plane, ii) movement
only in the heading direction ($\myss{^{n}}{_{t}}{\psi}=\myss{_{F}^{n}}{_{t}}{\chi}$).
If airship $n$ moves at velocity $\left\Vert \myss{_{F}^{n}}{_{t}}{\bm{v}}\right\Vert >v_{\min}$,
while the subject is not moving relative to the fluid ($\left\Vert \myss{_{F}^{S}}{}{\bm{v}}\right\Vert =0$),
then the only stable solution to keep the subject centered in camera
view is i) to mount the camera perpendicular to the motion direction
with $\gamma=\pm\frac{\pi}{2}$ and ii) to orbit on a curved, tangential
path with constant radius 
\begin{equation}
    \myss{_{S}^{n}}{_{t}}{r}=\left\Vert \myss{_{S}^{n}}{_{t}}{\bm{p}}\right\Vert\mathrm{.}
\end{equation}

This satisfies the equation of circular motion
\begin{equation}
    \myss{_{S}^{n}}{_{t}}{r}=\frac{\left\Vert \myss{_{S}^{n}}{_{t}}{\bm{v}}\right\Vert }{\myss{^{n}}{_{t}}{\dot{\psi}}}\mathrm{,}\label{eq:5}
\end{equation}with tangential speed $\left\Vert \myss{_{S}^{n}}{_{t}}{\bm{v}}\right\Vert $
and rotation rate $\myss{^{n}}{_{t}}{\dot{\psi}}$. Since $\myss{^{n}}{_{t}}{\dot{\psi}}$
can be controlled, this can be done for arbitrary radii regardless
of velocity constraints.

If the subject is moving ($\left\Vert \myss{_{F}^{S}}{}{\bm{v}}\right\Vert \neq0$) (Fig.~\ref{fig:orbit} inset),
it is still possible to orbit around $S$ while keeping it centered
in the camera view. To illustrate this intuitively, we mirror and rotate the system until, without limiting generality, the subject is moving along the negative X axis with $\myss{^{n}}{_{t}}{\dot{\psi}}>0$
and $\gamma=\frac{\pi}{2}$. To keep the subject centered in the camera,
the speed of airship $n$ must change to match the projected subject velocity component perpendicular to the camera axis. Adding to the airships orbital speed $\myss{_{S}^{n}}{_{t}}{r}\myss{^{n}}{_{t}}{\dot{\psi}}$ we calculate
\begin{equation}
    \left\Vert \myss{_{F}^{n}}{_{t}}{\bm{v}}\right\Vert =\myss{_{S}^{n}}{_{t}}{r}\myss{^{n}}{_{t}}{\dot{\psi}}-\cos\left(\myss{^{n}}{_{t}}{\psi}\right)\left\Vert \myss{_{F}^{S}}{}{\bm{v}}\right\Vert\mathrm{.}\label{eq:4}
\end{equation}

Since by assumption (in this subsection only) the airship
can only fly in the direction it is pointing, the distance to the
subject, which is the current orbit radius, will change with
\begin{equation}
    \myss{_{S}^{n}}{_{t}}{\dot{r}}=\sin\left(\myss{^{n}}{_{t}}{\psi}\right)\left\Vert \myss{_{F}^{S}}{_{t}}{\bm{v}}\right\Vert \mathrm{.}
\end{equation}

By integration, we introduce constant $r_{0}$ as a base radius. Thus
$\myss{_{S}^{n}}{_{t}}{\bm{r}}$ can be given as 
\begin{equation}
    \myss{_{S}^{n}}{_{t}}{r}=r_{0}-\cos\left(\myss{^{n}}{_{t}}{\psi}\right)\frac{\left\Vert \myss{_{F}^{S}}{}{\bm{v}}\right\Vert }{\myss{^{n}}{_{t}}{\dot{\psi}}}\mathrm{.}\label{eq:r in circle}
\end{equation}

Inserting (\ref{eq:r in circle}) in (\ref{eq:4}), we get
\begin{equation}
    \left\Vert \myss{_{F}^{n}}{_{t}}{\bm{v}}\right\Vert =\myss{^{n}}{_{t}}{\dot{\psi}}r_{0}-2\cos\left(\myss{^{n}}{_{t}}{\psi}\right)\left\Vert \myss{_{F}^{S}}{}{\bm{v}}\right\Vert\mathrm{,}\label{eq:7}
\end{equation}from which we calculate minimal and maximal airspeed during
the orbit as
\begin{equation}
    \min\left(\left\Vert \myss{_{F}^{n}}{_{t}}{\bm{v}}\right\Vert \right)=\myss{^{n}}{_{t}}{\dot{\psi}}r_{0}-2\left\Vert \myss{_{F}^{S}}{}{\bm{v}}\right\Vert\label{eq:7-2}
\end{equation}and
\begin{equation}
    \max\left(\left\Vert \myss{_{F}^{n}}{_{t}}{\bm{v}}\right\Vert \right)=\myss{^{n}}{_{t}}{\dot{\psi}}r_{0}+2\left\Vert \myss{_{F}^{S}}{}{\bm{v}}\right\Vert\mathrm{,}\label{eq:7-2-1}
\end{equation}for a desired yaw rate $\myss{^{n}}{_{t}}{\dot{\psi}}$ and base radius $r_{0}$, i.e., the distance to the stationary subject or the average distance to a moving subject over one orbit.

Multiple airships in a formation, i.e., $N>1$, can maintain
the angular constraints (2nd constraint in Subsec. \ref{subsec:Problem-Statement})
only when $\myss{^{n}}{_{t}}{\dot{\psi}}$ is the same for all airships.
For physically achievable (\ref{eq:2}) orbits that satisfy all the required perception constraints (Subsec.~\ref{subsec:Problem-Statement}),
\begin{equation}
\min\left(\left\Vert \myss{_{F}^{n}}{_{t}}{\bm{v}}\right\Vert \right)\ge v_{\min} ~~~ \mathrm{\,and\,} ~~~ \max\left(\left\Vert \myss{_{F}^{n}}{_{t}}{\bm{v}}\right\Vert \right)\leq v_{\max}\mathrm{,}\label{eq:7-2-2}
\end{equation}must hold.

If two airships $m,n$ are opposed by $\pi$ on the same orbit, at $\myss{^{m}}{_{t}}{\psi}=0$ and $\myss{^{n}}{_{t}}{\psi}=\pi$, the worst case scenario is that the velocity of airship $m$, given by (\ref{eq:7-2}), approaches $v_{\min}$ and the velocity of airship $n$, given by (\ref{eq:7-2-1}), approaches $v_{\max}$, both simultaneously. Solving for the subject velocity using (\ref{eq:7-2}) and (\ref{eq:7-2-1}) at this point leads to

\begin{equation}
\left\Vert \myss{_{F}^{S}}{}{\bm{v}}\right\Vert = \frac{\left(v_{\max}-v_{\min}\right)}{4}\mathrm{.}\label{eq:minmaxproof0}
\end{equation}

Consequently, the maximum magnitude of the subject velocity for which the whole orbit satisfies all the required conditions (Subsec.~\ref{subsec:Problem-Statement}) is given as
\begin{equation}
\left\Vert \myss{_{F}^{S}}{}{\bm{v}}\right\Vert \leq \frac{\left(v_{\max}-v_{\min}\right)}{4}\mathrm{.}\label{eq:minmaxproof2}
\end{equation}

\subsubsection*{Feasibility under reversal\label{subsec:Stability-under-Reversal}}

\begin{figure}[t]
\centering{}\includegraphics[width=1\columnwidth,trim={0 1.5cm 0 0},clip]{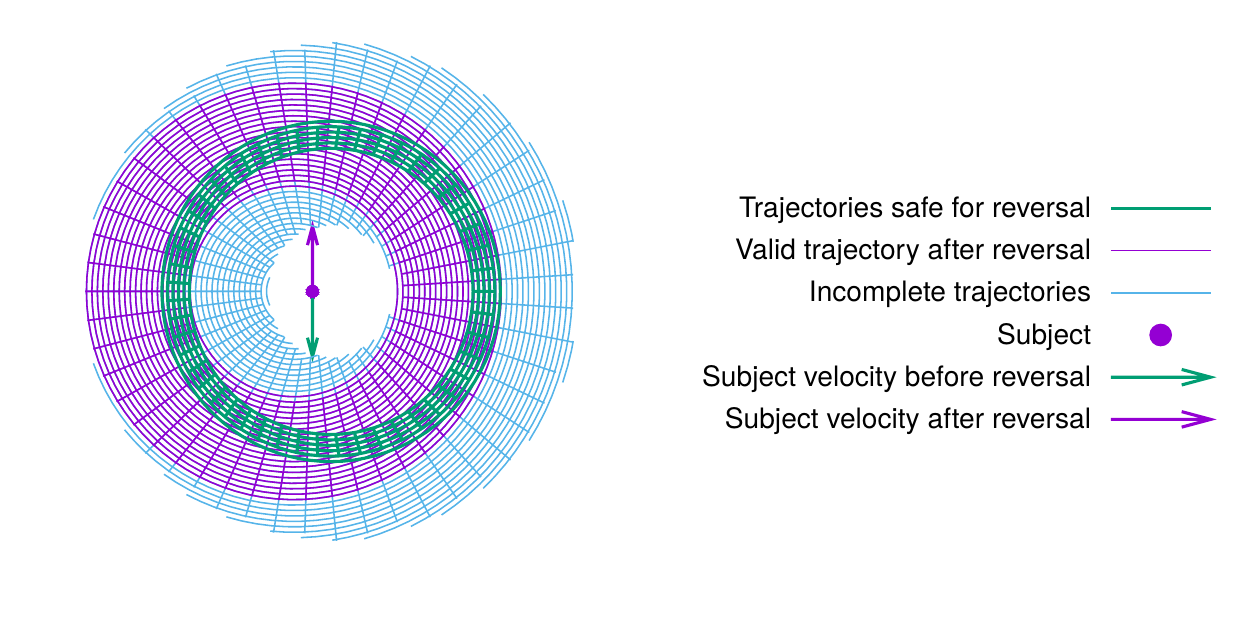}\caption{\label{fig:Reversible trajectories-1} Subject velocity reversal: Orbital space around moving subject for a fixed yaw rate
and airship velocity constraints $v_{\mathrm{min}}\leq\left\Vert \myss{_{F}^{n}}{}{\bm{v}}\right\Vert \leq v_{\mathrm{max}}$, with
visualization of the effect of reversal of $\myss{_{F}^{S}}{}{\bm{v}}$ direction.
For any $\myss{^{n}}{_{t}}{\dot{\psi}}$, the innermost valid trajectory
reaches $v_{\min}$ when flying opposite to the subject, the outermost
approaches $v_{\max}$ when flying parallel to the subject. A trajectory
is safe under reversal (green), if every point on the trajectory lies on a closed and valid purple trajectory.
Trajectories in blue can not be followed without violating constraints
(blank areas).}
\end{figure}
Although we consider the subject velocity as constant, we can also determine
robustness of this approach to changes in $\myss{_{F}^{S}}{}{\bm{v}}$.
For any given threshold $\myss{_{F}^{S}}{}{\bm{v}}$, the worst-case change is the sudden reversal of the motion direction, since this inverts the offset of the trajectory relative to the subject, as shown in Fig.~\ref{fig:Reversible trajectories-1}. The airship
might now be on an invalid (blue) orbit. Continuing on that trajectory
would then violate (\ref{eq:2}). However, a family of orbits exist, for which the whole
orbit, including minimum and maximum is always on a feasible trajectory
after reversal, as displayed in Fig. \ref{fig:Reversible trajectories-1}
in green. Analogous to (\ref{eq:minmaxproof2}), it can be shown that 
under single reversals safe trajectories exist if

\begin{equation}
    \left\Vert\myss{_{F}^{S}}{}{\bm{v}}\right\Vert\leq\frac{\left(v_{\max}-v_{\min}\right)}{8}\mathrm{.}
\end{equation}

In Summary, the analytic solution with the 2D assumptions for our
problem statement (Subsec. \ref{subsec:Problem-Statement}), is given
by (\ref{eq:r in circle}) subject to constraint (\ref{eq:minmaxproof2}).

\subsection{Airship Orbits in 3D with Realistic Physics\label{subsec:Steady-state physical system}}

A realistic airship model must take aerodynamic forces into account.
The hull of an airship, as depicted in Fig. \ref{fig:Accelerations-on-an}c
is subject to lift and drag forces \cite[pp 31-59]{von1963aerodynamics} and
can be modeled as an airfoil. Below we model the following physical parameters.

\subsubsection{Lateral angle of attack\label{subsec:Lateral-angle-of}}

In a real airship $\myss{^{n}}{_{t}}{\psi}\neq\myss{_{F}^{n}}{_{t}}{\chi}$
unless the airship flies in a straight line. To fly in a curve, a
centripetal force $F_{c}$ needs to act on the airship. This force
is a combination of the lateral engine thrust vector $F_{ty}$ and aerodynamic lift $F_{l}$ as shown in Fig. \ref{fig:Accelerations-on-an}c. Both force components are a function of the lateral angle of
attack $\myss{_{F}^{n}}{_{t}}{\beta}$, given as
\begin{equation}
\myss{_{F}^{n}}{_{t}}{\beta}=\myss{^{n}}{_{t}}{\psi}-\myss{_{F}^{n}}{_{t}}{\chi}\mathrm{.}
\end{equation}

\noindent The thrust component $F_{ty}$ is given as $F_{ty}=\tan\left(\myss{_{F}^{n}}{_{t}}{\beta}\right)F_{tx}$. If airspeed is maintained, $F_tx$ is proportional to $\left\Vert \myss{_{F}^{n}}{_{t}}{\bm{v}}\right\Vert^2$, Also, for small angles $\myss{_{F}^{n}}{_{t}}{\beta}\approx\tan{\myss{_{F}^{n}}{_{t}}{\beta}}$ and $F_{l}$ is proportional to both $\left\Vert \myss{_{F}^{n}}{_{t}}{\bm{v}}\right\Vert^2$ and $\myss{_{F}^{n}}{_{t}}{\beta}<\beta_\mathrm{stall}$ \cite[p. 44, Fig. 23]{von1963aerodynamics}. Therefore, we can combine both forces and approximate 

\begin{equation}
\myss{^{n}}{_{t}}{\bm{a}_c} \approx c_{l}  \left\Vert \myss{_{F}^{n}}{_{t}}{\bm{v}}\right\Vert^{2}\myss{_{F}^{n}}{_{t}}{\beta}  \mathrm{.}
\end{equation}
where $c_{l}$ is a system-specific coefficient, representing
the combined effect of the airship lift and drag. We will later enforce small $\myss{_{F}^{n}}{_{t}}{\beta}$ with control constraints $v_{\min}$ and $\dot{\psi}_{\max}$

As the airship rotates with $\myss{^{n}}{_{t}}{\dot{\psi}}$,
$\myss{_{F}^{n}}{_{t}}{\beta}$ increases until the lateral acceleration
approaches centripetal acceleration
\begin{equation}
\myss{^{n}}{_{t}}{\bm{a}_{c}}=\left\Vert \myss{_{F}^{n}}{_{t}}{\bm{v}}\right\Vert \myss{^{n}}{_{t}}{\dot{\psi}}\mathrm{.}
\end{equation}

At this point $\myss{_{F}^{n}}{_{t}}{\dot{\chi}}=\myss{^{n}}{_{t}}{\dot{\psi}}$
and $\myss{_{F}^{n}}{_{t}}{\dot{\beta}}=0$. Thus, for a fixed yaw
rate $\myss{^{n}}{_{t}}{\dot{\psi}}$ and for sufficient airspeed
$\left\Vert \myss{_{F}^{n}}{_{t}}{\bm{v}}\right\Vert \ge v_{\min}$ we
can approximate
\begin{equation}
\myss{_{F}^{n}}{_{t}}{\beta}\approx\frac{\myss{^{n}}{_{t}}{\dot{\psi}}}{c_{l}\left\Vert \myss{_{F}^{n}}{_{t}}{\bm{v}}\right\Vert }\mathrm{.}\label{eq:BETA}
\end{equation}

\subsubsection{Changes in velocity\label{subsec:Accelerated-system}}

If the airship is undergoing acceleration $\myss{_{F}^{n}}{}{a_{s}}$
in the direction of motion $\myss{_{F}^{n}}{_{t}}{\chi}$, then $\left\Vert \myss{_{F}^{n}}{_{t}}{\bm{v}}\right\Vert$,
given as 
\begin{equation}
\left\Vert \myss{_{F}^{n}}{_{t}}{\bm{v}}\right\Vert =\left\Vert \myss{_{F}^{n}}{_{t0}}{\bm{v}}\right\Vert +t\ \myss{_{F}^{n}}{}{a_{s}}\mathrm{.}\label{eq:derived absolute velocity}
\end{equation}

To calculate
\begin{equation}
\myss{_{F}^{n}}{_{t}}{\bm{v}}=\left\Vert \myss{_{F}^{n}}{_{t}}{\bm{v}}\right\Vert \left[\begin{array}{c}
\cos\left(\myss{_{F}^{n}}{_{t}}{\chi}\right)\\
\sin\left(\myss{_{F}^{n}}{_{t}}{\chi}\right)
\end{array}\right]\mathrm{,}\label{eq:derived velocity}
\end{equation}we determine 
\begin{equation}
\myss{_{F}^{n}}{_{t}}{\chi}=\myss{^{n}}{_{t}}{\psi}-\myss{_{F}^{n}}{_{t}}{\beta}\mathrm{,}\label{eq:26}
\end{equation}as a function of $\myss{^{n}}{_{t}}{\dot{\psi}}$ by plugging
(\ref{eq:BETA}) in (\ref{eq:26}) and obtain
\begin{equation}
    \myss{_{F}^{n}}{_{t}}{\chi}\approx\myss{^{n}}{_{t}}{\psi}-\frac{\myss{^{n}}{_{t}}{\dot{\psi}}}{c_{l}\left\Vert \myss{_{F}^{n}}{_{t}}{\bm{v}}\right\Vert }\mathrm{.}\label{eq:derived motion direction}
\end{equation}

\subsubsection{Roll angle}

Most airships are passively stable in roll and have their center of
mass underneath the center of lift and buoyancy. Consequently, they self-orient according to
the sum of gravity and centripetal acceleration. Although this
steady state will not be achieved immediately in dynamic situations, we can
approximate the roll angle of airship $n$ at time $t$ as
\begin{equation}
    \myss{^{n}}{_{t}}{\varphi}\approx\mathrm{atan}\left(\frac{\myss{_{F}^{n}}{_{t}}{\dot{\chi}}\left\Vert \myss{_{F}^{n}}{_{t}}{\bm{v}}\right\Vert }{g}\right)\mathrm{,}\label{eq:varphi-1}
\end{equation}
where $g\approx9.81\frac{\mathrm{m}}{\mathrm{s}\text{\texttwosuperior}}$
is earth gravity. In fixed wing aircraft the same formula describes
a \emph{coordinated turn} \cite[p. 153, Fig. 63]{von1963aerodynamics}.

\subsubsection{Altitude}

For altitude changes, we assume that the airship can change its vertical
speed arbitrarily and independently from its horizontal motion, within limits
\cite{Price:IAS:2021}. For the analytical solution, the necessary altitude
$\left(-\myss{_{S}^{n}}{_{t}}{p_{z}}\right)$ above $S$ to maintain
the subject centered in the camera view may be approximated based on radius
$\myss{_{S}^{n}}{_{t}}{r}$ and camera elevation $\phi$ (Fig. \ref{fig:Accelerations-on-an}b)
as
\begin{equation}
-\myss{_{S}^{n}}{_{t}}{p_{z}}\approx\frac{\tan\left(\myss{^{n}}{_{t}}{\varphi}-\phi\right)}{\myss{_{S}^{n}}{_{t}}{r}}\mathrm{,}\label{eq:altitude function}
\end{equation}

\noindent  however, this analytical solution ignores the effect of the airships pitch angle $\myss{^{n}}{_{t}}{\theta}$, which we explicitly consider in the numeric solution (Subsec. \ref{subsec:Numeric-motion-model}).

\subsubsection{Pitch angle}

Airships can change their altitude in a number of ways, not all of which require the same attitude changes. Yet, assuming that the nose points in the direction of travel is a good approximation in most cases. Therefore, pitch angle $\myss{^{n}}{_{t}}{\theta}$ is computed with
\begin{equation}
\myss{^{n}}{_{t}}{\theta}\approx\mathrm{atan}\left(\frac{-\myss{_{F}^{n}}{_{t}}{v_{z}}}{\myss{_{F}^{n}}{_{t}}{v_{h}}}\right)\mathrm{.}\label{eq:theta}
\end{equation}

In summary, one approximate analytical solution for airship orbits in 3D for our
problem statement (Subsec. \ref{subsec:Problem-Statement} ) can be
obtained by combining (\ref{eq:r in circle}), (\ref{eq:varphi-1})
and (\ref{eq:altitude function}), however the effect of changes in
$\myss{^{n}}{_{t}}{\theta}$ is not reflected in this analytical solution.
In the numeric solution (Subsec. \ref{subsec:Numeric-motion-model}
), we empirically show that changes in $\myss{^{n}}{_{t}}{\theta}$
make the orbits less eccentric than in (\ref{eq:r in circle}).

\subsection{Optimization Formulation for Numeric Solution\label{subsec:Numeric-motion-model}}

We can find an optimal trajectory for a formation starting state $\boldsymbol{X}_{0}$, by minimizing a cost function $C$ given as
\begin{equation}
C=\sum_{t,n}\left(k_{c}\myss{^{n}}{_{t}}{E_{c}}+k_{f}\myss{^{n}}{_{t}}{E_{f}}\right),\label{eq:Main Cost Function}
\end{equation}subject to constraints
\begin{equation}
v_{\min}\leq\left\Vert \myss{_{F}^{n}}{_{t}}{\bm{v}}\right\Vert \leq v_{\max} ~~~ \textrm{and} ~~~ \left\Vert \myss{_{F}^{n}}{_{t}}{v_{z}}\right\Vert \leq {v_{z\max}}\mathrm{.}
\end{equation}

$k_{c}\myss{^{n}}{_{t}}{E_{c}}$ keeps the subject $S$ centered in the camera images, while $k_{f}\myss{^{n}}{_{t}}{E_{f}}$ maintains angular separation between the airships around $S$. $k_{f}$ and $k_{c}$ can be chosen to prioritize these goals during transition periods. This does not change the converged orbit, as both cost terms are near $0$ for a fully converged trajectory. Here 
\begin{equation}
\myss{^{n}}{_{t}}{E_{c}}=\left\Vert \left[ {\begin{array}{ccc}
	k_{d} & 0 & 0 \\
    0 & 1 & 0 \\
    0 & 0 & 1 \\
\end{array} }\right]\left(\left[\begin{array}{c}
d_{c}\\
0\\
0
\end{array}\right]-\myss{^{n}}{_{t}}{\boldsymbol{R}}\left(-\myss{_{S}^{n}}{_{t}}{\bm{p}}\right)\right)\right\Vert ^{2}\mathrm{,}\label{Camera Pointing Cost}
\end{equation}

$\myss{^{n}}{_{t}}{\boldsymbol{R}}$ is the rotation matrix
combining $\bm{\Gamma}$ and $\myss{^{n}}{_{t}}{\bm{\Theta}}$ for a projection
of $S$ into the camera coordinate system of airship $n$, $k_{d}$
is a weight for the distance term, and $d_{c}$ is the optimal camera
distance. As shown in \cite{8784232}, a formation is optimal for
minimizing joint uncertainty in state estimate when equally spaced
for $3$ or more vehicles, but at $90\text{\textdegree}$ to each
other for $2$ vehicles. This is achieved with

\begin{equation}
\myss{^{n}}{_{t}}{E_{f}}=\begin{cases}
\underset{m\neq n}{\stackrel[m=1]{N}{\sum}}\left(\frac{\pi}{2}-\mathrm{acos}\left(\left|-\myss{_{S}^{n}}{_{t}}{\bm{p}}\right|\cdot\left|-\myss{_{S}^{m}}{_{t}}{\bm{p}}\right|\right)\right)^{2} & \mathrm{for\ }N=2\\
\underset{m\neq n}{\stackrel[m=1]{N}{\sum}}\max\left(0,\frac{2\pi}{N}-\mathrm{acos}\left(\left|-\myss{_{S}^{n}}{_{t}}{\bm{p}}\right|\cdot\left|-\myss{_{S}^{m}}{_{t}}{\bm{p}}\right|\right)\right)^{2} & \mathrm{for\ }N\ge3\mathrm{.}
\end{cases}\label{Formation Cost}
\end{equation}For $N\ge3$, $\myss{^{n}}{_{t}}{E_{f}}$ penalizes airships that are separated by less than the desired angular distance for an evenly spaced formation $\frac{2\pi}{N}$. This has a repulsive effect and optimizes for equal spacing without enforcing any specific order.

We solve for control input $\boldsymbol{U}$ at time $t$,
consisting of vehicle controls $\myss{^{n}}{_{t}}{\bm{u}}$, with
\begin{equation}
\boldsymbol{U}_{t}=\left[\myss{^{1}}{_{t}}{\bm{u}},\ldots,\myss{^{N}}{_{t}}{\bm{u}}\right] ~~~ \textrm{and} ~~~ \myss{^{n}}{_{t}}{\bm{u}}=\left[\myss{^{n}}{_{t}}{\dot{\psi}},\myss{^{n}}{_{t}}{\dot{v}_{h}},\myss{^{n}}{_{t}}{\dot{v}_{z}}\right]\mathrm{.}
\end{equation}

\begin{figure*}
\centering

\begin{subfigure}[b]{.45\textwidth}
\centering
\includegraphics[width=1\columnwidth,trim={0 0.6cm 0 0},clip]{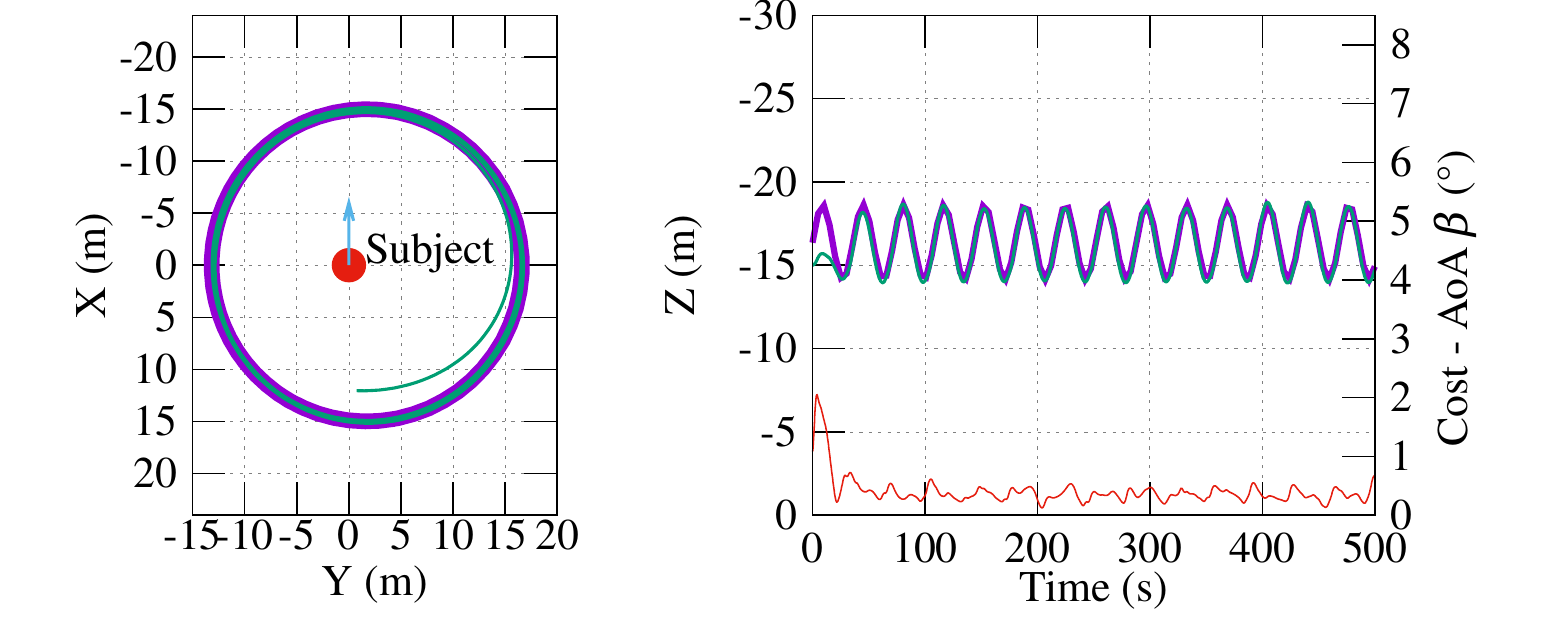}
\vspace{-2\baselineskip}
\caption{~~~~~~~~~~~}
\end{subfigure}
\begin{subfigure}[b]{.45\textwidth}
\centering
\includegraphics[width=1\columnwidth,trim={0 0.6cm 0 0},clip]{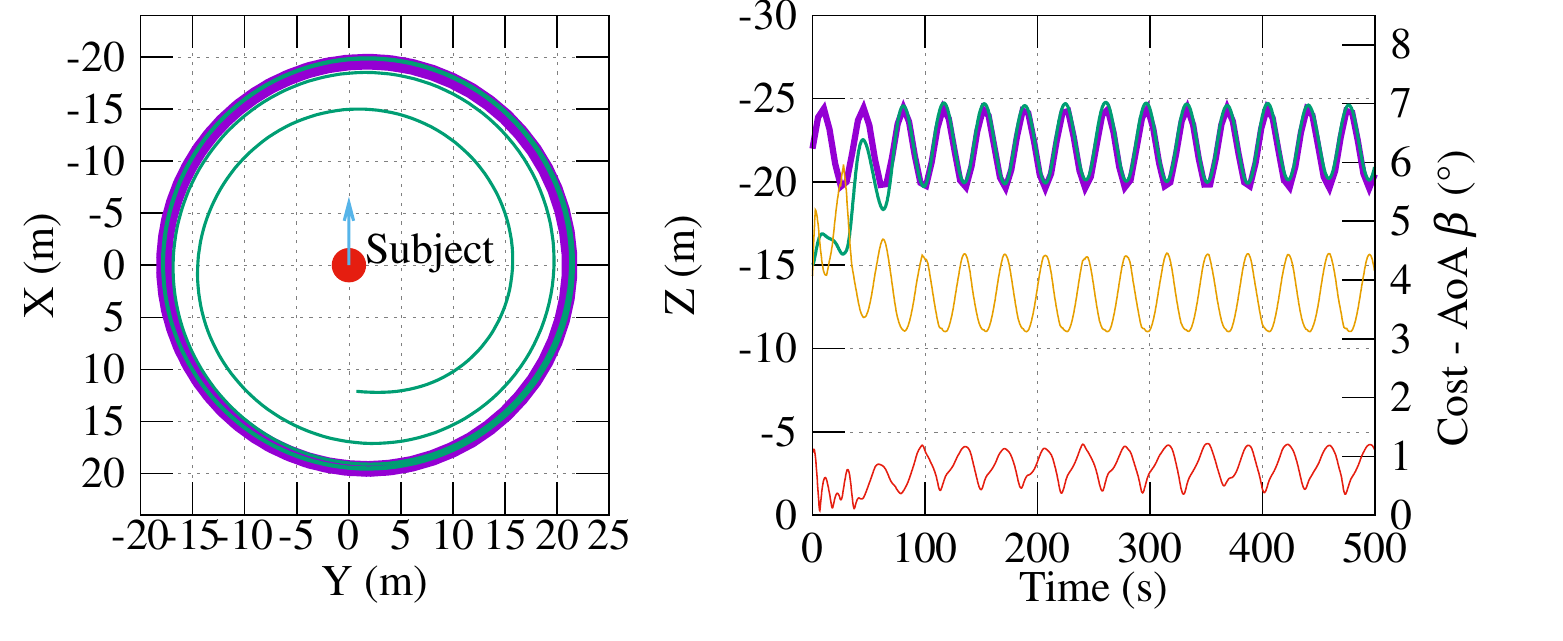}
\vspace{-2\baselineskip}
\caption{~~~~~~~~~~~}
\end{subfigure}\quad

\vspace*{15pt}

\begin{subfigure}[b]{.45\textwidth}
\centering
\includegraphics[width=1\columnwidth,trim={0 0.6cm 0 0},clip]{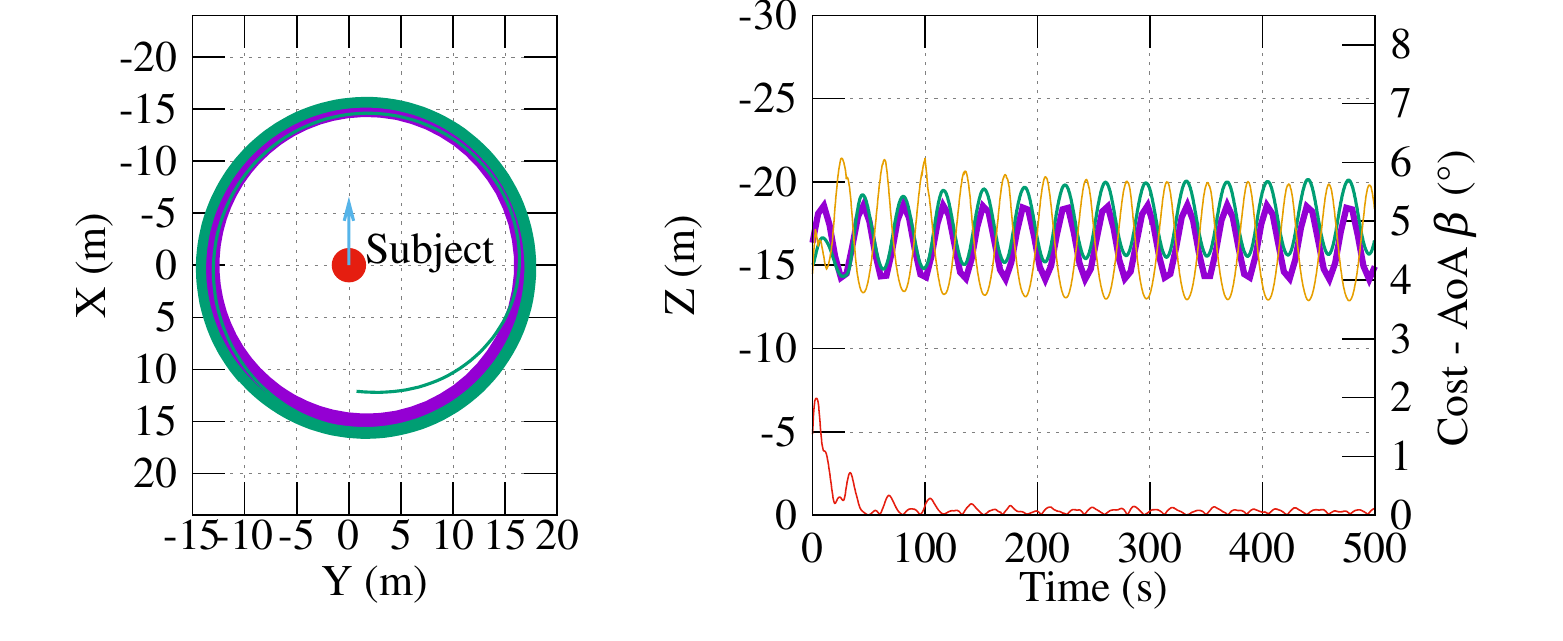}
\vspace{-2\baselineskip}
\caption{~~~~~~~~~~~}
\end{subfigure}
\begin{subfigure}[b]{.45\textwidth}
\centering
\centering{}\includegraphics[width=1\columnwidth,trim={0 0.6cm 0 0},clip]{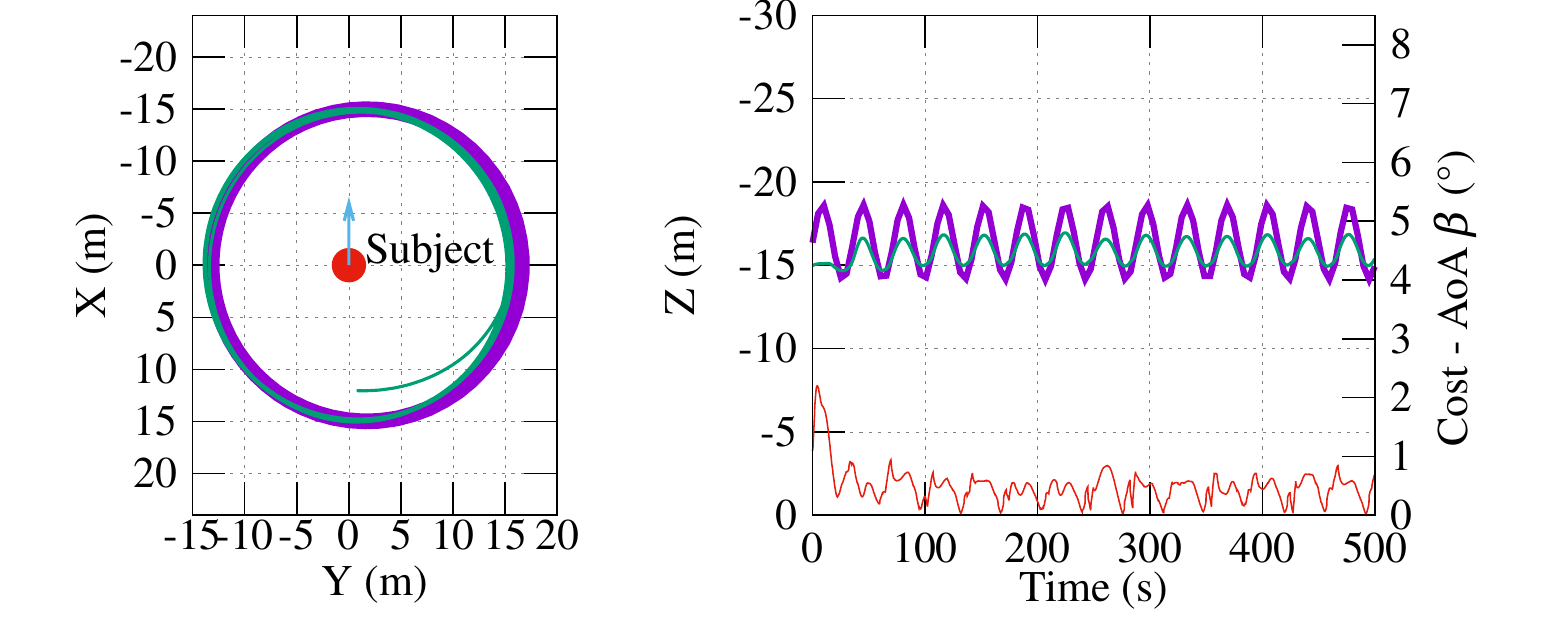}
\vspace{-2\baselineskip}
\caption{~~~~~~~~~~~}
\end{subfigure}

\vspace*{10pt}

\begin{subfigure}[b]{1\textwidth}
\centering
\includegraphics[width=1\columnwidth, scale=0.8]{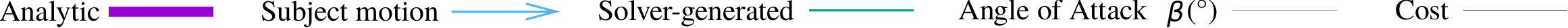}
\end{subfigure}

\caption{
Orbital space around moving subject for a fixed yaw rate
$\myss{^{n}}{_{t}}{\dot{\psi}}=\mathrm{const}$ and velocity constraints on
the airship $v_{\min}\leq\left\Vert
\myss{_{F}^{n}}{_{t}}{\bm{v}}\right\Vert \leq v_{\max}$.  The analytic
solution or closest approximation is shown in purple. The green trajectory
is a solver generated numeric solution, attempting to keep the camera
centered on the subject. The red line is cost $\myss{^{n}}{_{t}}{E_{c}}$.
\textbf{a)} \label{fig:Orbital-space-around}2D case. We assume
$c_{l}\rightarrow\infty$, which implies
$\myss{^{n}}{_{t}}{\psi}=\myss{_{F}^{n}}{_{t}}{\chi}$
and$\myss{_{F}^{n}}{_{t}}{\beta}=\myss{^{n}}{_{t}}{\theta}=0$. The solver
generated trajectory and the analytic trajectory are a close match.
\textbf{b)} \label{fig:Beta_not_zero}, \textbf{c)} \label{fig:Beta_not_zero_2} Analysis of the effect of
$\myss{_{F}^{n}}{_{t}}{\beta}$ for $\gamma=\frac{\pi}{2}$ (c))
respectively $\gamma=87\text{\textdegree}$. $\myss{^{n}}{_{t}}{\theta}$ is
assumed zero.  At a sensor angle of $\gamma=\frac{\pi}{2}$ this angular offset
forces the airship on an outward spiral. Once the maximum
constraint-satisfying distance is reached (purple), the airship can no
longer maintain the subject centered in the camera view, as seen in the
non-zero cost (red). At a sensor angle of $\gamma=87\text{\textdegree}$
the solver converges to a solution only slightly further out than
$\min\left(\left\Vert \myss{_{F}^{n}}{}{\bm{v}}\right\Vert
\right)=v_{\mathrm{min}}$ (purple), at which the effects of AoA
$\myss{_{F}^{n}}{_{t}}{\beta}$ are fully compensated over a full orbit,
allowing the subject to remain centered in the camera image within tight
tolerances.
\textbf{d)} \label{fig:Theta_not_zero} Analysis of the effect of $\myss{^{n}}{_{t}}{\theta}$. AoA $\myss{_{F}^{n}}{_{t}}{\beta}$
is assumed $0$. The numeric solution (green) takes $\myss{^{n}}{_{t}}{\theta}$
into account, the analytic solution does not. The resulting optimal trajectory has less altitude variation than the
analytic solution, but shares the same minimum radius. However, the maximum
radius is significantly reduced, implying that there could
be viable orbits in 3D with $\left\Vert\myss{_{F}^{S}}{}{\bm{v}}\right\Vert >\frac{1}{4}\left(v_{\max}-v_{\min}\right)$.
}
\end{figure*}

\begin{figure}
\centering{}\includegraphics[width=1\linewidth,trim={0.5cm 3cm 1.5cm 0},clip]{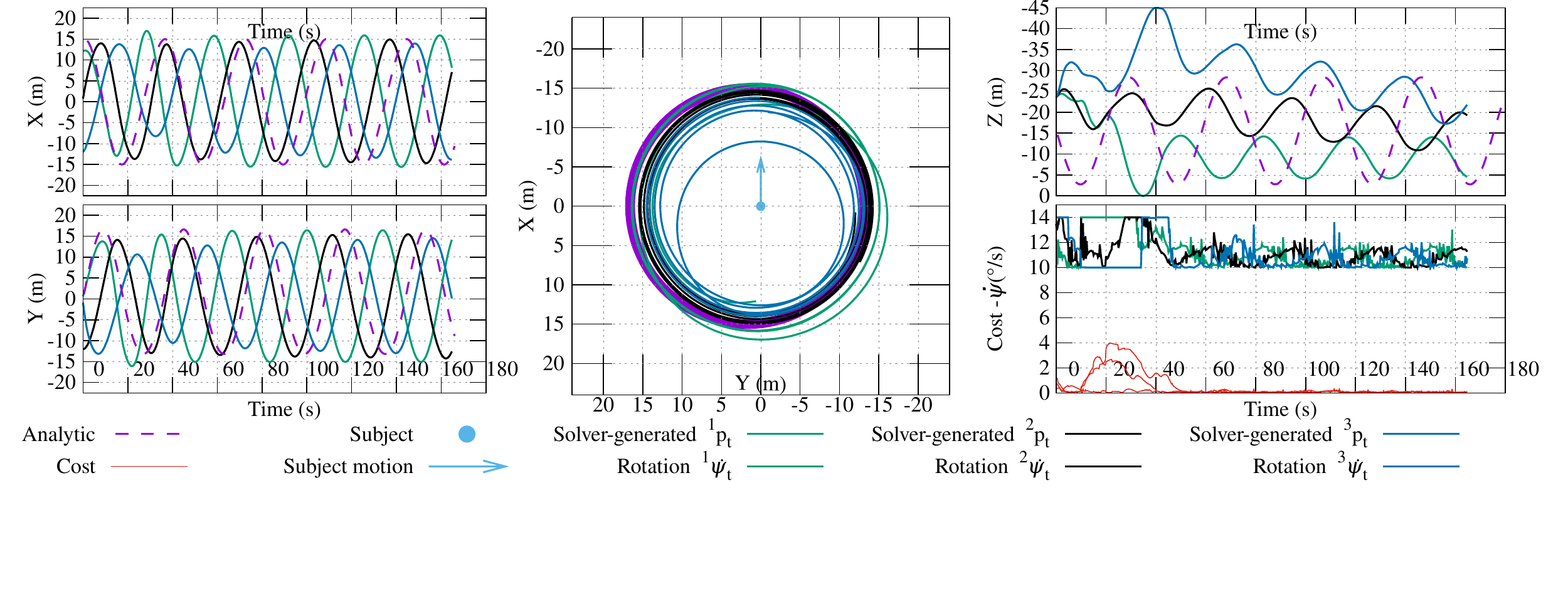}\caption{\label{fig:3vehicleFomration}Numerically solved optimal formation
of 3 airships around a moving subject. Trajectories have a non-optimal
starting point, leading to some non-optimal subject centering in order
to converge on a global optimal formation within the first $1\frac{1}{2}$
orbits. The red plot shows only the cost term $\myss{^{n}}{_{t}}{E_{c}}$.}
\end{figure}

We define the discrete time state transition consisting
of subject state transition and state transitions for all airships,
with
\begin{equation}
\boldsymbol{X}_{t+\Delta t}=\left[\myss{^{S}}{_{t+\Delta t}}{\bm{x}},\myss{^{1}}{_{t+\Delta t}}{\bm{x}},\ldots,\myss{^{N}}{_{t+\Delta t}}{\bm{x}}\right]^{\top}\mathrm{.}
\end{equation}

The subject state transition is given as
\begin{equation}
    \myss{^{S}}{_{t+\Delta t}}{\bm{x}}=\left[\myss{^{S}}{_{t}}{\bm{p}}+\Delta t\myss{^{S}}{}{\bm{v}}\right]\mathrm{.}
\end{equation}

The airship state transition is given by
\begin{equation}
\myss{^{n}}{_{t+\Delta t}}{\bm{x}}=\left[
\myss{^{n}}{_{t+\Delta t}}{\bm{p}}, 
\myss{^{n}}{_{t+\Delta t}}{\bm{v}}, 
\myss{^{n}}{_{t+\Delta t}}{\bm{\Theta}}\right]^\top\mathrm{,}
\end{equation}where $\myss{^{n}}{_{t+\Delta t}}{\bm{p}}$ is calculated as
\begin{equation}
\myss{^{n}}{_{t+\Delta t}}{p}=\myss{^{n}}{_{t}}{\bm{p}}+{\left(\myss{^{n}}{_{t}}{\bm{v}}\right)}\Delta t\mathrm{.}\label{eq: new position}
\end{equation}$\myss{^{n}}{_{t+\Delta t}}{\bm{v}}$ is calculated as
\begin{equation}
    \myss{^{n}}{_{t+\Delta t}}{v}=\myss{_{F}^{n}}{_{t+\Delta t}}{\bm{v}}+\myss{^{F}}{}{\bm{v}}\mathrm{,}\label{Velocity-1}
\end{equation}where
\begin{equation}
\myss{_{F}^{n}}{_{t+\Delta t}}{\bm{v}}=\left[\begin{array}{c}
\left(\myss{_{F}^{n}}{_{t}}{v_{h}}+\myss{_{F}^{n}}{_{t}}{\dot{v}_{h}}\Delta t\right)\cos\left(\myss{_{F}^{n}}{_{t+\Delta t}}{\chi}\right)\\
\left(\myss{_{F}^{n}}{_{t}}{v_{h}}+\myss{_{F}^{n}}{_{t}}{\dot{v}_{h}}\Delta t\right)\sin\left(\myss{_{F}^{n}}{_{t+\Delta t}}{\chi}\right)\\
\myss{_{F}^{n}}{_{t}}{v_{z}}+\myss{_{F}^{n}}{_{t}}{\dot{v}_{z}}\Delta t
\end{array}\right]\mathrm{.}\label{Velocity}
\end{equation} $\myss{^{n}}{_{t+\Delta t}}{\bm{\Theta}}$ is computed as 
\begin{equation}
\myss{^{n}}{_{t+\Delta t}}{\bm{\Theta}}=\left[\begin{array}{c}
\myss{^{n}}{_{_{t+\Delta t}}}{\varphi}\\
\myss{^{n}}{_{_{t+\Delta t}}}{\theta}\\
\myss{^{n}}{_{t}}{\psi}+\myss{^{n}}{_{t}}{\dot{\psi}}\Delta t
\end{array}\right]\mathrm{.}\label{eq:Formula_For_Theta}
\end{equation}

\noindent In (\ref{eq:Formula_For_Theta}), $\myss{_{F}^{n}}{_{t+\Delta t}}{\chi}$,$\myss{^{n}}{_{t+\Delta t}}{\varphi}$
and $\myss{^{n}}{_{t+\Delta t}}{\theta}$ are calculated according
to (\ref{eq:derived motion direction}), (\ref{eq:varphi-1}) and
(\ref{eq:theta}), respectively.

\subsection{Model Evaluation\label{subsec:Model-Evaluation}}

We implemented the optimization problem described above (Subsec.~\ref{subsec:Numeric-motion-model}) using OpEn/PANOC \cite{open2020,panoc2017}.
Additional common cost terms, omitted for brevity, are implemented to minimize control effort and add soft constraints around hard state constraint boundaries.
To increase the accuracy at large $\Delta t$, we use  4th order Runge
Kutta to integrate over $\intop\myss{_{F}^{n}}{_{t}}{\bm{v}}\delta t$
to propagate $\myss{^{n}}{_{t}}{\bm{p}}$ in (\ref{eq: new position}).

A single airship trajectory, optimized using the model in Sec.~\ref{subsec:Numeric-motion-model}
converges to the analytical optimal solution described in Sec.~\ref{subsec:Analytic-solution-for}
as shown in Fig.~\ref{fig:Orbital-space-around}a), if $\myss{_{F}^{n}}{_{t}}{\beta}$
and $\myss{^{n}}{_{t}}{\theta}$ are set to $0$ and $\myss{^{n}}{_{t}}{\dot{\psi}}$
is set constant. If the starting state is not on the optimal
trajectory, there is a short transitional phase, where the cost is
minimal but not zero. $k_{d}$ is set to $0$ for this evaluation
to compute the optimal trajectory, without regard to the distance
between the airship and subject $S$.

As shown in Fig. \ref{fig:Beta_not_zero}b), a non-zero lateral angle
of attack $\myss{_{F}^{n}}{_{t}}{\beta}$ results in a trajectory,
where the subject is not centered in the camera, but is offset
by approximately $\myss{_{F}^{n}}{_{t}}{\beta}$ . $\myss{_{F}^{n}}{_{t}}{\beta}$
is smaller at higher airspeed, as such the numeric solution approaches
the largest radius possible within constraints. To compensate for this
effect, one solution is to adjust the camera azimuth $\gamma$ (Fig.~\ref{fig:Accelerations-on-an}). An intuitive correction would be
$\gamma=\frac{\pi}{2}-\mathrm{mean}\left(\myss{_{F}^{n}}{_{t}}{\beta}\right)$,
so that the average would be zero. This mean depends not only on the model
parameters but also on the mean velocity, which in turn depends on
the orbit radius. We can estimate it by applying (\ref{eq:BETA})
on a circular orbit with the same parameters. In Fig. \ref{fig:Beta_not_zero}b),
$\myss{_{F}^{n}}{_{t}}{\beta}$ oscillates between $3{^\circ}$ and
$4.5{^\circ}$.

Figure \ref{fig:Beta_not_zero_2}c) shows that setting $\gamma=87\text{\textdegree}$
solves this problem and allows an optimal solution close to the minimum
radius. Intuitively, a larger orbit results in higher velocities and
a smaller $\myss{_{F}^{n}}{_{t}}{\beta}$. When $\gamma>\frac{\pi}{2}-\mathrm{mean}\left(\myss{_{F}^{n}}{_{t}}{\beta}\right)$,
the airship needs to then fly an inward spiral to keep the subject
centered in view. Similarly, a too tight orbit will lead to a larger
lateral angle of attack with $\gamma<\frac{\pi}{2}-\mathrm{mean}\left(\myss{_{F}^{n}}{_{t}}{\beta}\right)$.
Consequently, the formation will naturally converge on an orbit with
$\gamma=\frac{\pi}{2}-\mathrm{mean}\left(\myss{_{F}^{n}}{_{t}}{\beta}\right)$.
This is not a limitation for our approach, as we still have $\myss{^{n}}{_{t}}{\dot{\psi}}$
as a controllable variable to change the desired distance to the subject.

Figure \ref{fig:Theta_not_zero}d) shows the effect of $\myss{^{n}}{_{t}}{\theta}$
with a camera angled $\phi=-45\text{\textdegree}$ downwards. Intuitively,
when the nose of the airship raises in a climb, which is the case
whenever the radius $\myss{_{S}^{n}}{_{t}}{r}$ grows (\ref{eq:altitude function}), the camera looks further ahead. Compensating this to keep subject
$S$ in view, enforces a sharper turn, diminishing the $\myss{_{S}^{n}}{_{t}}{r}$
increase. Symmetrically, when descending during shrinking $\myss{_{S}^{n}}{_{t}}{r}$,
the camera looks behind. Both effects reduce the change in $\myss{_{S}^{n}}{_{t}}{r}$.
As the change in radius $\myss{_{S}^{n}}{_{t}}{r}$ is now less pronounced
as in the analytic 2D case in Subsec. \ref{subsec:Analytic-solution-for},
the upper bound for the subject velocity (\ref{eq:minmaxproof2})
becomes relaxed.

As shown before, solutions to the 2D simplified orbiting problem are guaranteed to exist if $\left\Vert\myss{_{F}^{S}}{}{\bm{v}}\right\Vert \leq\frac{1}{4}\left(v_{\max}-v_{\min}\right)$. However, with the change in $\myss{^{n}}{_{t}}{\theta}$, solutions might exist for $\left\Vert\myss{_{F}^{S}}{}{\bm{v}}\right\Vert > \frac{1}{4}\left(v_{\max}-v_{\min}\right)$, but we have no analytical solution for the exact boundary, which depends
on parameters $\gamma$, $\phi$, $\myss{_{F}^{n}}{_{t}}{\beta},$ 
$c_{l}$, $\myss{^{n}}{_{t}}{\varphi}$ and $\myss{^{n}}{_{t}}{\dot{\psi}}$.

Finally we apply the solver to compute optimal formation trajectories
for $3$ airships. In order to allow airships to adjust their relative
angles to subject $S$, we no longer fix the yaw rate $\myss{^{n}}{_{t}}{\dot{\psi}}$,
but allow the solver to adjust $\myss{^{n}}{_{t}}{\dot{\psi}}$ within
new constraints $\dot{\psi}_{\min}\leq\myss{^{n}}{_{t}}{\dot{\psi}}\leq\dot{\psi}_{\max}$
to converge to an optimal multi airship formation, shown in Fig. \ref{fig:3vehicleFomration}.
The constraints are identical for all airships. As cost weight $k_{f}>0$
is now in effect, the optimal solution is a trade-off between $\myss{^{n}}{_{t}}{E_{c}}$
and $\myss{^{n}}{_{t}}{E_{f}}$.

\section{Implementation, Experiments and Results\label{sec:Experiments}}
\setlength{\belowcaptionskip}{-5pt}

\subsection{Implementation and Experimental Setup\label{subsec:ROS-MPC-Implementation}}

We implement the model in Sec. \ref{subsec:Numeric-motion-model} as
a nonlinear MPC for real airships.  Like \cite{8784232}, it is computationally
centralized, but distributed from an information perspective as each airship
runs its own instance of the formation controller based on the information it
receives from other airships. We reduce the planning
horizon to approximately $\frac{1}{3}$ orbit.
This planning horizon is long enough
to conduct a $180{^\circ}$ turn if necessary, e.g., in case of sub-optimal
starting conditions or to avoid collisions. Additional control constraints
are enforced to limit the change in $\myss{^{n}}{_{t}}{\dot{\psi}}$
in each timestep, and state constraints such as minimum-height
to avoid ground contact. We also added additional non-convex constraints
to enforce sufficient distances between airships at all times and
prevent collisions between them and possible static obstacles. In
the absence of static obstacles, any solution close to the optimum
has the airships maximally spread out due to cost term (\ref{Formation Cost}),
which naturally forms a large convex corridor that optimal orbits
reside in. 

We modify the open source airship controller in \cite{Price:IAS:2021}
to take $\myss{^{n}}{_{t}}{\dot{\psi}}$, $\myss{_{F}^{n}}{_{t}}{\dot{v}_{h}}$
and $\myss{_{F}^{n}}{_{t}}{\dot{v}_{z}}$ as control inputs directly.
We can measure $\myss{_{F}^{n}}{}{v_{h}}$ using a pitot probe and
$\myss{^{n}}{}{\bm{x}}_t$ and $\myss{^{n}}{_{t}}{\dot{\psi}}$ using GPS
and inertial sensors, which allows us to estimate wind vector $\myss{^{F}}{_{t}}{\bm{v}}$
by inverting (\ref{Velocity}) with a numerical approximation method.
We employ the cooperative perception framework of AirCap \cite{8394622}
to estimate $\myss{^{S}}{_{t}}{\bm{x}}$ from $\myss{^{1\ldots N}}{_{t}}{\bm{x}}$
and visual detection in the camera images using a single shot detector
neural network \cite{10.1007/978-3-319-46448-0_2} and a distributed
Bayesian filter. We assume $\myss{^{F}}{}{\bm{v}}$ and $\myss{^{S}}{}{\bm{v}}$
constant over the planning horizon. We estimate the model specific
parameters $c_{l}$ for real airships by measuring $\myss{_{F}^{n}}{_{t}}{\beta}$
during a constant rate turn (\ref{eq:BETA}). With this, we can use
our MPC to control a real or simulated airship. With a horizon of
$10$ timesteps and $\Delta t=1.25\mathrm{s}$, the OpEN engine typically
converges within $\sim20\mathrm{ms}$ on a Jetson TX2 embedded
computer, which is sufficiently fast for real time application.

For all experiments, we set $k_{c}=1$, $k_{d}=0.6$, $d_{c}=15\mathrm{m}$,
$k_{f}=100$ as optimization weights. Camera and vehicle parameters
for the simulated airships were $\gamma=82\text{\textdegree}$, $\phi=-30\text{\textdegree}$,
$c_{l}=0.24$. The state constraints were set to $v_{\min}=0.5\frac{\mathrm{m}}{\mathrm{s}}$,
$v_{\max}=4.0\frac{\mathrm{m}}{\mathrm{s}}$, $v_{z\max}=0.5\frac{\mathrm{m}}{\mathrm{s}}$,
$\dot{\psi}_{\min\brokenvert\max}=\pm18\frac{\text{\textdegree}}{\mathrm{s}}$,
allowing both left and right turns.

\subsection{Experiment Description and Evaluation Metrics\label{subsec:Experimental-Setup}}

Experiments 1-3 are done in simulation. In Experiment 1, we evaluate different formation sizes up to $N=6$
airships around a stationary subject, with constant wind of $\left\Vert\myss{^{F}}{}{\bm{v}}\right\Vert=0.6\frac{\mathrm{m}}{\mathrm{s}}$.
In Experiment 2, we investigate the effect of wind and wind gusts
with $N=3$ airships. In Experiment 3, we investigate the effect of
subject motion, including sudden turns, with and without wind. Finally,
in Experiment 4, we evaluate our MPC in an outdoor environment with
a real airship but a simulated stationary subject.

In experiments 1–3, we evaluate 4 different metrics. 

\begin{enumerate}[leftmargin=*]
\item The visual tracking accuracy, which measures how well the estimated subject position,
given as $\myss{^{S}}{_{t}}{\bm{p}}$, agrees with the ground-truth $\myss{_{\mathrm{GT}}^{S}}{_{t}}{\bm{p}}$  in simulation, calculated as $\left\Vert\myss{^{S}}{_{t}}{\bm{p}}-\myss{_{\mathrm{GT}}^{S}}{_{t}}{\bm{p}}\right\Vert $.
\item Tracking uncertainty, measured as the trace of the covariance matrix of the subject position estimate.
\item Subject visibility, which is the percentage of video frames that
have the subject in the field of view (FOV). The subject is considered
in FOV if its projected center point is distanced less than half the
camera resolution from the image center point. For this, the camera
is simulated as a pinhole projection with $640\times480$ pixels resolution
and $90{^\circ}$ camera FOV.
\item The proximity to the image center, measured as the distance in pixels of the subject center to the camera image center. 
\end{enumerate}

In experiment 4, since no tracking takes place, we use only 3) and
4), but in addition we compare whether the actually
flown trajectory to simulated trajectories under similar wind conditions.

\subsection{Experiment 1 - Formation Size\label{subsec:Experiment-1}}

\begin{figure}
\centering{}\includegraphics[width=1\columnwidth]{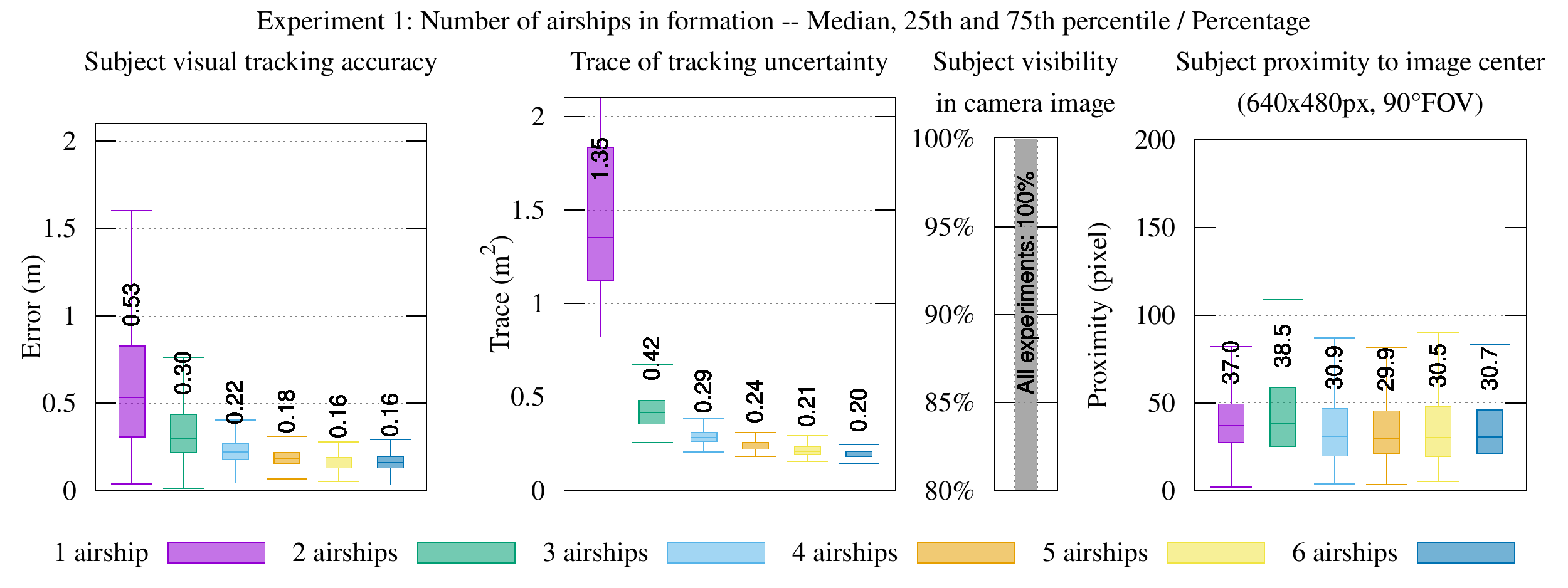}
\caption{\label{fig: Experiment1}Experiment 1: Variation of Formation Size.}
\end{figure}

Figure \ref{fig: Experiment1} summarizes the effect of formation size
on the tracking formation with $N=1,2,3,4,5 ~ \mathrm{ and } ~ 6$. All evaluations were
conducted with wind speed $\left\Vert \myss{^{F}}{}{\bm{v}}\right\Vert =0.6\frac{\mathrm{m}}{\mathrm{s}}$.
For $N=1$ there is a larger tracking error and uncertainty, which is expected
since a single airship cannot reliably estimate the depth from a single camera
image. Consistent with \cite{8394622,8784232}, for $N>=2$ the tracking error
and uncertainty becomes smaller with each additional airship in the formation.
All formations managed to maintain sight of the subject $100\mathrm{\%}$
of the time, with the distance to the camera center averaging around
$30\mathrm{px}$ and never exceeding $40\mathrm{px}$.

\subsection{Experiment 2 - Wind Velocity\label{subsec:Experiment-2}}

\begin{figure}
\centering{}\includegraphics[width=1\columnwidth]{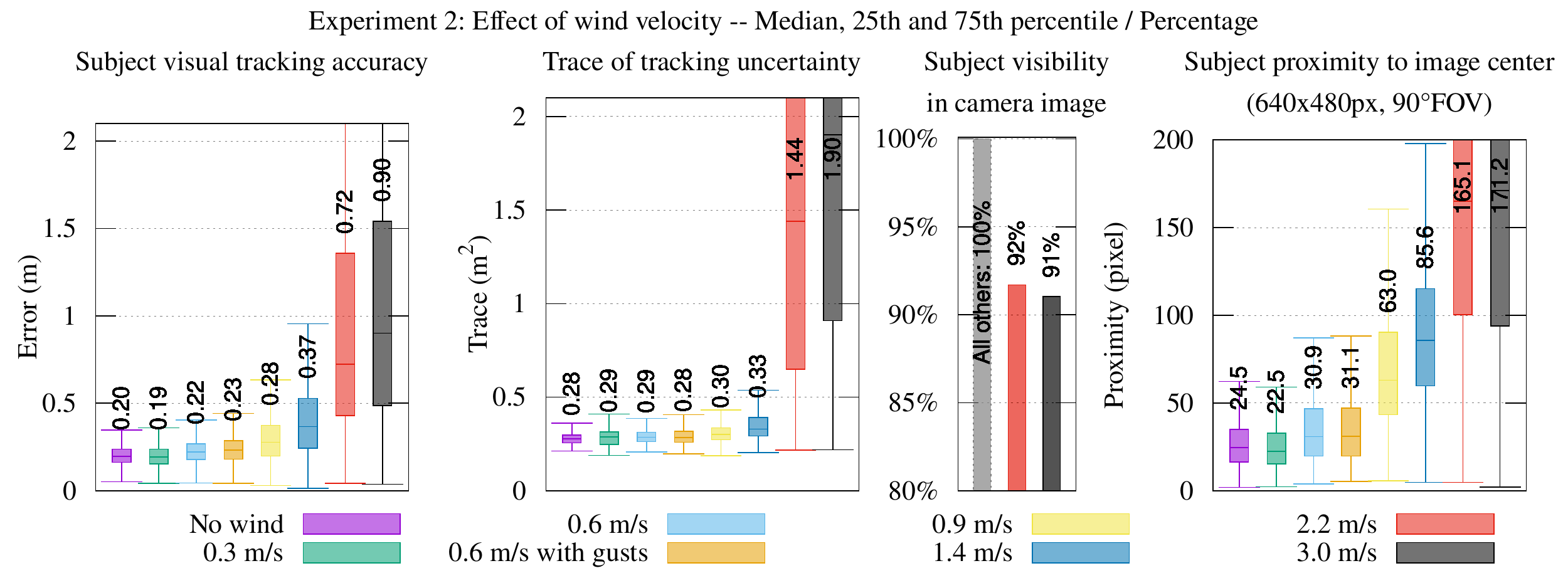}
\caption{\label{fig: Experiment2}Experiment 2: Variation of Wind Velocity.}
\end{figure}
Figure \ref{fig: Experiment2} shows the effect of wind on the tracking
formation with $\left\Vert \myss{^{F}}{}{\bm{v}}\right\Vert
=0\frac{\mathrm{m}}{\mathrm{s}},0.3\frac{\mathrm{m}}{\mathrm{s}},0.6\frac{\mathrm{m}}{\mathrm{s}},0.9\frac{\mathrm{m}}{\mathrm{s}},1.4\frac{\mathrm{m}}{\mathrm{s}},2.2\frac{\mathrm{m}}{\mathrm{s}}$
and $3.0\frac{\mathrm{m}}{\mathrm{s}}$. All evaluations were conducted with
$N=3$ airships and a stationary subject. For wind gusts we utilize the Dryden turbulence model at turbulence level ``moderate''.
Up to $1.4\frac{\mathrm{m}}{\mathrm{s}}$ wind, there is a small but consistent
decrease in tracking accuracy, however as the wind exceeds
$\frac{v_{\max}-v_{\min}}{4}$ (from $\left\Vert \myss{^{F}}{}{\bm{v}}\right\Vert = 0.9 \frac{\mathrm{m}}{\mathrm{s}}$ ) the MPC no longer manages to keep subject $S$
well centered in camera images which is consistent with our analysis on 2D orbits. With $\left\Vert \myss{^{F}}{}{\bm{v}}\right\Vert
>=2.2\frac{\mathrm{m}}{\mathrm{s}}$ the airships could no longer maintain sight
of the subject through the entire orbit and with stronger wind there were
increasingly large blind sectors during which $S$ could not be observed. This
also results in increased tracking error, although the subject was still
visible in the cameras for close to $90\mathrm{\%}$ of all camera frames.

\subsection{Experiment 3 - Subject Not Stationary\label{subsec:Experiment-3}}
\setlength{\belowcaptionskip}{-15pt}
\begin{figure}
\centering{}\includegraphics[width=1\columnwidth]{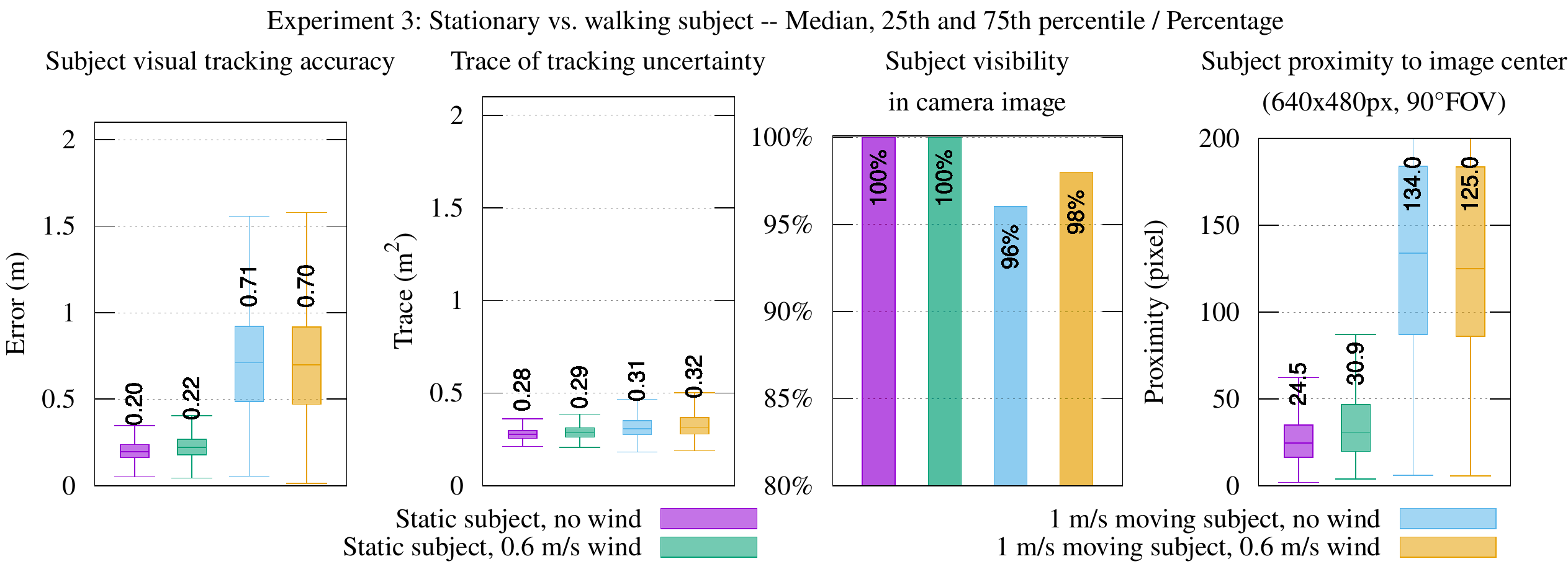}
\caption{\label{fig: Experiment3}Experiment 3: Subject not Stationary}
\end{figure}
Figure \ref{fig: Experiment3} shows the effect of subject motion.  For two of
the trials, with no wind and $0.6\frac{\mathrm{m}}{\mathrm{s}}$ wind, the
subject is stationary, while for the third and fourth, the subject follows a
repeated predefined trajectory. This subject path consists of straight and
curved segments with sharp direction changes.  This simulates the worst case
described in Subsec. \ref{subsec:Stability-under-Reversal}.  In the trials with
moving subject, the subject velocity is $\sim1\frac{\mathrm{m}}{\mathrm{s}}$,
which is much larger than $\frac{v_{\max}-v_{\min}}{8}$. Consequently, keeping
the subject centered through the reversals is not possible.  However, the MPC
still maintains the subject in the camera FOV for $96\mathrm{\%}$ and
$98\mathrm{\%}$ of all video frames in those trials, with or without
wind. Typically only one out of three airships loses sight at a time and only
for a very short time whenever an unanticipated subject direction change
happens.

\subsection{Experiment 4 - Real World Experiment\label{subsec:Experiment-4}}

\begin{figure}
\centering{}\includegraphics[width=1\columnwidth,clip]{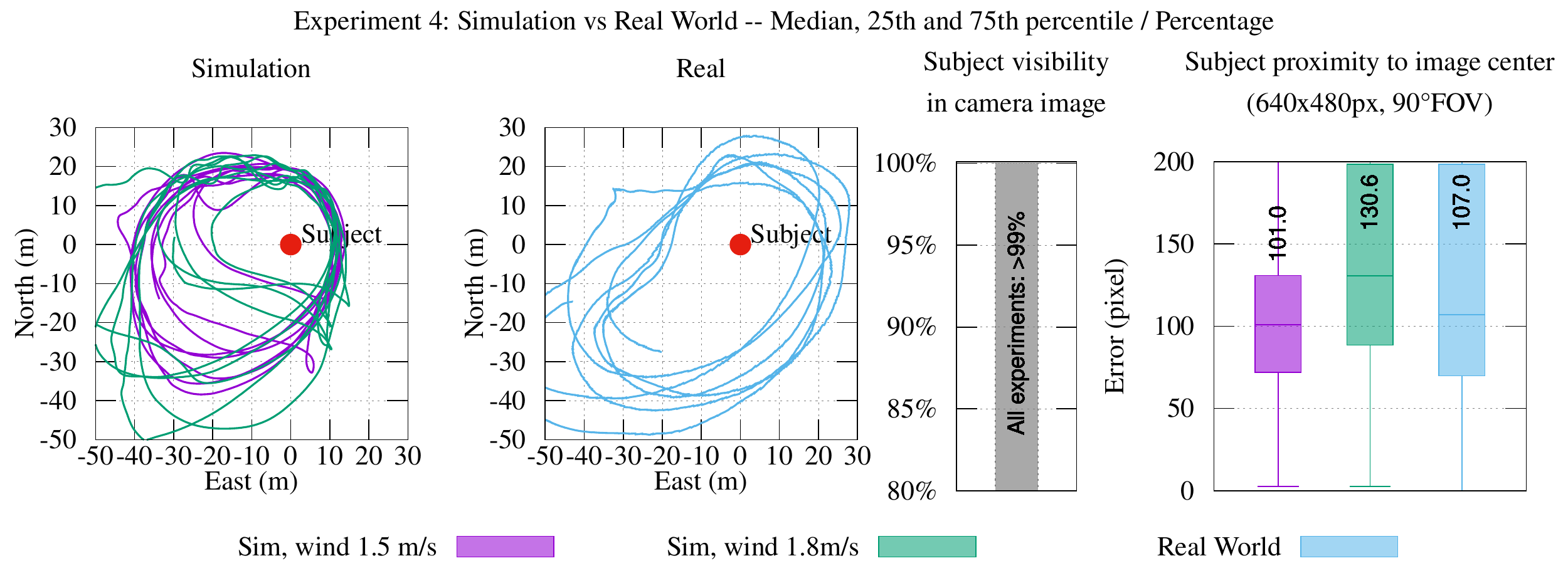}\caption{\label{fig:Real-world-experiment}Experiment 4: Real world flight.
Comparison of real world flight results to two simulated flights in similar wind conditions with $1.5$ and $1.8$ m/s average wind speeds, yielding comparable results.}
\end{figure}

We conducted a real world experiment with one airship to verify the
applicability of our control method. To isolate evaluation of the control
method, no tracking took place, the subject position was fixed at $\left[ 0,0,0
\right]$, providing ground-truth for the subject location. The wind was measured
between $1$ and $2$ m/s with gusts in excess of $4$ m/s from East to North-East.
($\myss{{}^{S}_{F}}{}{\bm{v}} > v_{\max}$).  Under these conditions, stable equally
spaced formations with multiple airships would not have been possible. In
Fig.~\ref{fig:Real-world-experiment}, we show the real world flight alongside
two simulated flights at different mean wind speeds using the Dryden model with
``severe'' simulated turbulence intensity. Both the resulting trajectories and
performance of the controller w.r.t. keeping the camera pointed at the subject in the real experiment
are in line with the simulation results under equivalent wind conditions.

\section{Conclusion and Outlook\label{sec:Summary}}

In this article, we presented an MPC-based method for formation control of
airships with frame-fixed cameras. The goal of the MPC is to keep a moving
subject within the field of view of the cameras of all the airships
simultaneously, while adhering to other motion and view constraints. We derived
analytical solutions in a simplified 2D case for boundary conditions on the
subject velocity. We also approximated analytical solutions for the realistic 3D
case where possible, excluding the effect of the pitch angle, and averaging
over the lateral angle of attack. We formulated our objective as an
optimization problem and solved it using a numerical solver. This was followed
by an MPC implementation for real-time application. Through extensive
simulations, we show our method's efficacy, accuracy and reliability. Through a
real world experiment, we demonstrate its speed, stability and real world
applicability. In future work, we will investigate the problem of collision and
obstacle avoidance within the context of  airship formation control.
Generalization to other non-holonomic vehicles, e.g., fixed-wing aircraft is
also planned.

 \bibliographystyle{IEEEtran}
\phantomsection\addcontentsline{toc}{section}{\refname}\bibliography{paper}

\end{document}